\documentclass[9pt]{article}
\usepackage{spconf,amsmath,graphicx}

\usepackage{cleveref}
\usepackage{helvet}
\usepackage{tabularx}
\usepackage{enumitem}
\usepackage{mdwlist}
\usepackage{algorithm}
\usepackage{algpseudocode}
\usepackage{amsthm}
\makeatletter
\def\thm@space@setup{%
  \thm@preskip=1pt
  \thm@postskip=\thm@preskip 
}
\makeatother

\usepackage{hhline}
\usepackage{amssymb}
\usepackage{lipsum}
\usepackage{balance,setspace}
\usepackage{multirow,comment, color}
\usepackage{amsmath}
\usepackage{soul}
\usepackage{arydshln}
\usepackage{xcolor}

\newlength\myindent
\def\Width{0\kern2\tabcolsep\ldots\kern1\tabcolsep0}
\usepackage{etoolbox}
\newcommand{\zerodisplayskips}{%
  \setlength{\abovedisplayskip}{3pt}
  \setlength{\belowdisplayskip}{3pt}
  \setlength{\abovedisplayshortskip}{3pt}
  \setlength{\belowdisplayshortskip}{3pt}}
\appto{\normalsize}{\zerodisplayskips}
\appto{\small}{\zerodisplayskips}
\appto{\footnotesize}{\zerodisplayskips}

\def\D{{\mathbf D}}

\def\A{{\mathbf A}}

\definecolor{green}{rgb}{0, 0, 1}
\title{Exploiting Low-Dimensional Structures to Enhance DNN Based Acoustic Modeling in Speech Recognition} 
\name{Pranay Dighe$^{\star \circ}$ \qquad Gil Luyet$^{\dagger \star}$ \qquad Afsaneh Asaei$^{\star}$ \qquad Herv\'{e} Bourlard$^{\star \circ}$}

\address{$^{\star}$Idiap Research Institute, Martigny, Switzerland\\
$^{\circ}$\'{E}cole Polytechnique F\'{e}d\'{e}rale de Lausanne (EPFL), Switzerland\\
$^{\dagger}$University of Fribourg, Switzerland\\
{\small \tt \{pranay.dighe,afsaneh.asaei,herve.bourlard\}@idiap.ch, gil.luyet@unifr.ch} 
}

\begin{document}
\ninept

\maketitle

\begin{abstract} 
We propose to model the acoustic space of deep neural network (DNN) class-conditional posterior probabilities as a union of low-dimensional subspaces. To that end, the training posteriors are used for dictionary learning and sparse coding. Sparse representation of the test posteriors using this dictionary enables projection to the space of training data.   
Relying on the fact that the intrinsic dimensions of the posterior subspaces are indeed very small and the matrix of all posteriors belonging to a class has a very low rank, we demonstrate how low-dimensional structures enable further enhancement of the posteriors and rectify the spurious errors due to mismatch conditions. 
The enhanced acoustic modeling method leads to improvements in continuous speech recognition task using hybrid DNN-HMM (hidden Markov model) framework in both clean and noisy conditions, where upto $15.4\%$ relative reduction in word error rate (WER) is achieved.

\end{abstract}
\vspace{-1mm}
\begin{keywords}
Sparse coding, Dictionary learning, Deep neural network, Union of Low Dimensional Subspaces, Acoustic modeling.
\end{keywords}

\vspace{-2mm}
\section{Introduction}\label{sec:intro}
\vspace{-1mm}
A need for sparse representations for better acoustic modeling of speech has been advocated consistently for better characterization of the underlying low-dimensional and parsimonious structure of speech~\cite{bilmes2006hmms,bengio2009learning,sainath2011exemplar,saon2012bayesian}. 
Two major emerging trends, namely deep neural networks (DNN) and exemplar-based sparse modeling, are different approaches of exploiting sparsity in speech representations to achieve invariance, discrimination and noise separation~\cite{deng2013machine,saon2012bayesian,gemmeke2008noise}. 

On the other hand, speech utterances are formed as a union of words which in turn consist of phonetic components and sub-phonetic attributes. 
Each linguistic component is produced through activation of a few highly constrained articulatory mechanisms leading to generation of speech data in union of low-dimensional subspaces~\cite{deng2004switching,king2007speech,lee2001functional}.   
However, most existing speech classification and acoustic modeling methods do not explicitly take into account the multi-subspace structure of the data.

The present study focuses on exploiting the multi-subspace low-dimensional structure of speech learned from the training data to enhance DNN based acoustic modeling of unseen test data. Hence, this also has the potential to enable domain adaptation and handling mismatch in the framework of DNN based acoustic modeling.


\vspace{-2mm}
\subsection{Prior Works}\label{sec:prior}
\vspace{-1mm}
Sparse representation has been proven powerful as features used for acoustic modeling.   
As argued in \cite{bengio2009learning}, if data is projected into high-dimensional space, the underlying structures are dis-entangled. These structures form a union of low-dimensional subspaces which models the non-linear manifold where speech data resides. 
Prior work on sparse representation includes exemplar-based methods~\cite{sainath2011exemplar, gemmeke2011exemplar} where sparse representation, learned using spectral features achieve promising performance in automatic speech recognition (ASR) specially due to their robustness in handling noise and corruption. 

Recent advancement in DNN based acoustic modeling relies on estimation of highly sparse sub-word class-conditional posterior probabilities. While the conventional Gaussian mixture models (GMM) are statistically inefficient in modeling data lying on or near non-linear manifolds~\cite{hinton2012deep,deng2004switching,king2007speech}, DNNs achieve accurate sparse acoustic modeling through multiple layers of non-linear transformations \cite{yu2013feature}. The hidden layers of DNN successively learn underlying structures at different levels and express them as highly invariant and discriminative representations towards deeper layers. While enforcing sparsity constraints during DNN training is mostly employed for the purpose of regularization to prevent overfitting, various studies have shown that sparsity in DNN architectures directly contributes towards simpler networks and superior performance in ASR. Successful application of sparse activity \cite{liu2015neuron} (very few neurons being active), sparse connectivity \cite{yu2012exploiting} (very few non-zero weights) as well as better performance of sparsity inducing techniques like dropout neural network training \cite{srivastava2014dropout} confirm the belief that \textit{`sparser'} is better for acoustic modeling in ASR.
\vspace{-2mm}
\subsection{Motivation and Contributions}\label{sec:motivation}
\vspace{-1mm}
We point out two issues with respect to the state-of-the-art DNN based acoustic models which motivate further consideration of sparse modeling: 
\begin{description}
\item{Q1.} Previous studies~\cite{tasha2015exploring,mohamed2012understanding} have found sparse activations in DNNs by showing how individual neurons in hidden layers learn being selectively active in different ways towards distinct phone patterns. Since this sparsification learned by hidden layers is not explicitly hand-crafted, we ask upto what extent the union of low-dimensional subspaces structure for speech is actually being exploited by DNNs ?
\item{Q2. } Despite of being effective in seen conditions, DNNs are found highly sensitive to unseen variations in data~\cite{yu2013feature}. The mismatch condition causes erroneous estimates of posterior probabilities which is exhibited as spurious noises in the output posterior probabilities. Can we correct these errors through a low-dimensional model to improve acoustic modeling in noisy conditions ?
\end{description}
In this paper, we address those issues by explicit modeling of the underlying structures in speech using prior knowledge that speech data lives in the union of low-dimensional subspaces. We implement this idea using the principled dictionary learning and sparse coding algorithms over DNN posterior probabilities to recover sparse representations where non-zero values correspond to class-specific subspaces. 
These subspace sparse representations are then used to \textit{enhance} the original DNN posterior probabilities through dictionary based reconstruction. We build upon compressive sensing and subspace sparse recovery theory to provide theoretical support for validity of our approach. We also elaborate on our choice of features (DNN based posterior probabilities) and algorithms for dictionary learning~\cite{mairal2010online} and structured sparse coding~\cite{sprechmann2011c} which essentially distinguish our approach from previous exemplar based sparse representation methods~\cite{sainath2010sparse, sainath2011exemplar, gemmeke2011exemplar}. We demonstrate improvements in performance achieved by the proposed enhanced acoustic modeling in hybrid DNN-HMM continuous ASR system using Numbers'95 database \cite{Cole95newtelephone} and show increased robustness in noisy conditions. 

In the rest of the paper, the proposed subspace sparse acoustic modeling method is elaborated in Section~\ref{sec:union_of_subspace}. The experimental analysis are carried out in Section~\ref{sec:analysis}. Section \ref{sec:conclusions} provides the concluding remarks and directions for future work.





\vspace{-1mm}
\section{Subspace Sparse Acoustic Modeling}\label{sec:union_of_subspace}
\vspace{-1mm}
In this section, we model the space of DNN class-conditional posterior probabilities as a union of low-dimensional subspaces. Relying on the subspace sparse representation, we show how the posteriors can be enhanced for more accurate class-specific representations.
\vspace{-3mm}
\subsection{Subspace Sparse Representation}
\vspace{-2mm}
Speech features reside on or near non-linear manifolds which can be best characterized by union of low-dimensional subspaces. The proposed approach relies on the fact that a data point in a union of subspaces can be more efficiently reconstructed using a sparse combination of data points from its own subspace than data points from other subspaces, thus  resulting in a \textit{subspace-sparse representation}~\cite{elhamifar2013sparse}.

To state it more precisely, let $\mathbf{S}=\{\mathcal{S}_\ell\}^L_{\ell=1}$ be a set of linear disjoint subspaces associated to $L$ classes in $\mathbb{R}^m$ such that the dimensions of individual subspaces $\{r_\ell\}_{\ell=1}^L$ are smaller than the dimension of the actual space, i.e. $\forall\ell\textrm{, }r_\ell < m$. Speech features $z$ lie in the union $\cup_{\ell=1}^{L} \mathcal{S}_\ell$ of these low-dimensional subspaces. Let $\mathbf{D}_\ell \in \mathbb{R}^{m\times n_\ell}$ be the class-specific over-complete dictionary  for subspace $\mathcal{S}_\ell$ where $n_\ell$ is the number of atoms in $\D_\ell$ and $n_\ell>r_\ell$. Each data point in $\mathcal{S}_\ell$ can then be represented as a sparse linear combination of the atoms from $\D_{\ell}$. 

Defining $\ell_1$-norm of a vector (denoted by $\|.\|_1$) as the sum of the absolute values of its components, the \textit{subspace sparse recovery} (SSR) property~\cite{elhamifar2013sparse} for union of disjoint subspaces asserts that $\ell_1$-norm sparse representation of a data point over collection of all class-specific dictionaries $\{\D_{\ell}\}_{\ell=1}^L$ can lead to separation of the class-specific subspaces by selecting atoms only from the underlying class of the data point for its reconstruction. Thus, the obtained sparse representations have activations only for the atoms corresponding to the actual subspace $\mathcal{S}_\ell$ where $z$ lives.

Considering a speech utterance as the union of words, phones or sub-phonetic components, the subspaces $\mathcal{S}_\ell$ can be modeled at different levels (time granularity) corresponding to any of these speech units. Consequently a dictionary $\D$ can be constructed by learning basis sets $\D_\ell$ for individual classes. In the present study, we focus on context-dependent senones (c.f. Section~\ref{sec:features}) for their superior quality in DNN-HMM framework. Nevertheless there is no theoretical/algorithmic impediment in applying it for larger units such as words.

The rigorous proof of SSR property (see Theorem 2 in~\cite{elhamifar2013sparse}) requires certain conditions and assumptions on disjoint subspaces. Since we train DNN with binary senone target outputs, the intersection of senone subspaces is expected to be a rare event and suggests disjointedness of subspaces. Although further theoretical  analysis is beyond the scope of the present work, experiments conducted in Section~\ref{sec:analysis} empirically confirm that SSR property indeed holds for subspace-sparse modeling of senones.


\vspace{-2mm}
\subsection{Class-Specific Dictionary Learning}\label{sec:sparse}
There are two key considerations for dictionary learning in sparse subspace acoustic modeling. Namely, the choice of features and algorithmic developments. 
\vspace{-3mm}
\subsubsection{Senone Posterior Probabilities as Speech Features}\label{sec:features}
\vspace{-1.5mm}
A posterior feature $z$ is a vector consisting of class-conditional probabilities at the output layer of DNN. In contrast to spectral features, posterior features are proven highly effective for sparse modeling~\cite{dighe2015sparse,asaei2010analysis}. They are inherently sparse and invariant to speaker/environmental conditions presented in the DNN training data. Although we choose to work with posterior probabilities at context-dependent senone levels (tied triphone states)~\cite{yu2015automatic}, the theoretical underpinning of the proposed approach is applicable to any type of speech units. 
\vspace{-3mm}
\subsubsection{Dictionary Learning and Sparse Coding Algorithms}\label{sec:algorithms}
\vspace{-1.5mm}
Building on our previous work on dictionary learning for sparse modeling of posterior features~\cite{dighe2015sparse}, we use the online dictionary learning~\cite{mairal2010online} algorithm for solving $l_1$ sparse coding problem expressed as
\begin{equation}\small \label{eq:optimize_func}
\arg \min_{\substack \D,\A} \sum_{t=1}^{T} \| z_t - \D\,\alpha_{t} \|_2^2 + \lambda{\|\alpha_{t}\|}_1, \textrm{  s.t.  } \| d_j \|_2^2\leq1\, \forall{j}
\end{equation}
where $\A=[\alpha_1 \hdots \alpha_T]$ and $d_j$ denotes each atom of the dictionary. 

Class-specific data of senone posterior features is obtained through GMM-HMM based forced alignment on training data, which is then used to learn individual over-complete basis set $\D_{\ell}$ for each senone subspace $S_{\ell}$ using dictionary learning algorithm. These class-specific dictionaries are concatenated into a larger dictionary $\D=[\mathbf{D}_{1} \cdots \mathbf{D}_{\ell} \cdots \mathbf{D}_{L}]$ for subspace-sparse acoustic modeling. Since any posterior feature obtained from DNN lies in a union of subspaces $\cup_{\ell=1}^{L}\mathcal{S}_\ell$, a test posterior feature $z$ can be reconstructed using the atoms of dictionary $\D$. According to SSR property, only the atoms associated to the correct class (underlying subspace) of $z$ will be used for sparse representation. 

It may be noted that dictionary learning approach is fundamentally different from dictionary construction using a random subset \cite{sainath2011exemplar, gemmeke2011exemplar} of training features since we use all of the training data to compute an over-complete basis set for sparse representation which is far smaller (less than 3\% in case of Numbers'95 database) than the actual collection size yet more effective in sparse representation~\cite{dighe2015sparse}.  
\vspace{-3mm}
\subsection{Enhanced Acoustic Modeling}\label{sec:sparse_representations}
\vspace{-2mm}
We use group sparsity based hierarchical Lasso algorithm~\cite{sprechmann2011c} for sparse coding to enforce group sparsity in $\alpha$ based on the internal partitioning of dictionary $\D$ into senone-specific sub-dictionaries $\D_{\ell}$. The high dimensional group sparse representation $\alpha$ is computed for each DNN output posterior feature $z$ by sparse recovery over $\D$. Projection of a test posterior feature $z$ on training data space is given by computing $\D\alpha$.

\begin{figure}[t]
  \centering
  \includegraphics[width=0.97\columnwidth]{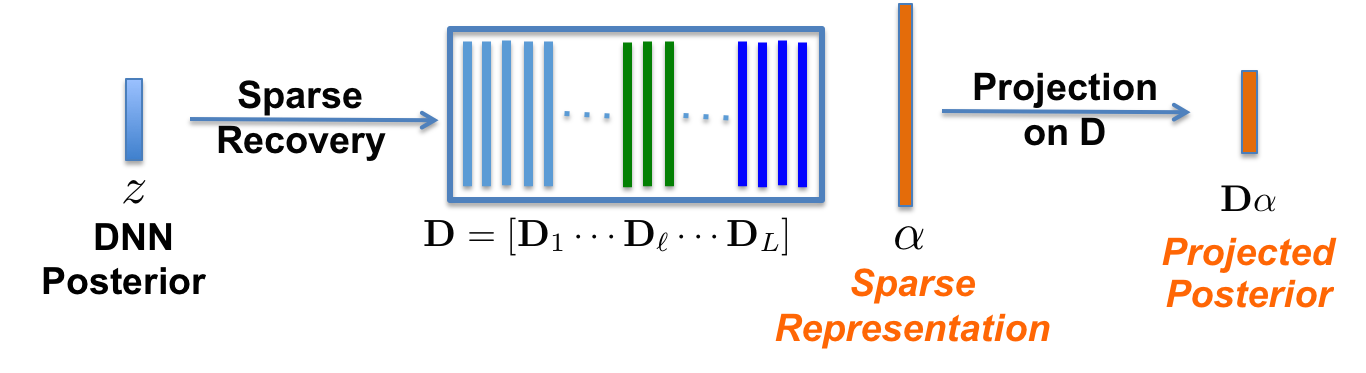}
  \caption{\small DNN output senone posteriors $z$ are projected to the space of training posteriors using $\D\alpha$. Resulting projected posteriors are used for typical decoding in DNN-HMM framework.}  
  \label{fig:dict_sparse}
\end{figure}

Note that $\D\alpha$ is an approximation of posterior feature $z$ based on $\ell_{1}$-norm sparse reconstruction using atoms of $\D$.  Consequently, it has the same dimension as $z$ and it is forced to lie in a probability simplex by normalization. Figure \ref{fig:dict_sparse} summarizes this procedure. 
\vspace{-3mm}
\section{Experimental Analysis}\label{sec:analysis}
\vspace{-2mm}
In this section, we provide empirical analysis of the theoretical results established in Section~\ref{sec:union_of_subspace}. These experiments confirm that the information bearing components of DNN class-conditional probabilities indeed live in a very low-dimensional space. Exploiting this structure enables enhancement of DNN based acoustic models and removes the effect of high-dimensional noise leading to improvement in DNN-HMM speech recognition performance. 
\vspace{-2mm}
\subsection{Database and Speech Features}\label{sec:regular}
We use Numbers'95 database for this study where only the utterances consisting of digits are considered (more details in~\cite{dighe2015sparse}). The phoneset includes 27 phones and accordingly 557 context dependent tied states referred to as senones are learned by forced alignment of the training data using Kaldi speech recognition toolkit~\cite{povey2011kaldi}.  
A DNN is trained using sequence discriminative training~\cite{vesely2013sequence} with 3 hidden layers each having 1024 nodes. For every 10 ms speech frame, the DNN input is a vector of MFCC+$\Delta$+$\Delta\Delta$ features with a context of 9 frames (39$\times$9=351 dimension). The DNN output is a vector of posterior probabilities corresponding to 557 senone classes. We use DNN posteriors as features $z$ for dictionary learning and sparse coding~\eqref{eq:optimize_func}.
\vspace{-2mm}
\subsection{Low-rank Posterior Reconstruction}\label{sec:uos-reconstruct}
As explained in Sections \ref{sec:sparse}--\ref{sec:sparse_representations}, DNN posteriors are used to learn senone-specific dictionaries $\D_{\ell}$ from the training data. Number of atoms $n_\ell$ in each senone dictionary $\D_\ell$ is approximately 100. A value of $\lambda=0.2$, optimized on development data, was used for sparse coding to get sparse representations $\alpha$. Subsequently $\D\alpha$ projected posterior probabilities are computed for the test data. Sparsity leads to selection of a few subspaces of the training data resulting in new test posteriors which (1) live in low-dimensions, (2) are projected onto the subspace of the training posteriors, and (3) separated from the subspaces of other senone classes. We investigate these properties below through further analysis. 

To provide an insight into the dimension of the senone subspaces, we construct matrices of 1000 class-specific senone posteriors and 
compute the number of singular values required to preserve 95\% variability of the data. Due to skewed distribution of the posteriors, we take their log prior to singular value decomposition. We refer to the number of required singular values as roughly the ``Rank'' of senone matrices.
An ideal posterior feature should have its maximum component at the support indicating its associated class. Hence, we group the posteriors as ``correct'' if the maximum component corresponds to the correct class and ``incorrect'' if the maximum component corresponds to the incorrect class. Table~\ref{table:analysis} shows the average number of required singular values over all senones for DNN and projected posteriors. Another approach referred to as robust PCA based posteriors will be discussed in the subsequent section.  
{\footnotesize
\begin{table}[b]\footnotesize
\centering
\begin{tabular}{lccc}
\hline
& DNN & Projected & Robust PCA \\
\hhline{====}
Rank-Correct   & 36.6 & 11.9 &  7.6 \\
Rank-Incorrect & 45.5 & 21.7 &  11.7 \\
\hline
\end{tabular}
\caption{\footnotesize Comparison of ``Rank'' of DNN posterior matrix, projected posterior matrix and RPCA senone posterior matrix.}
\label{table:analysis}
\end{table}}

We can see that the ``correct'' posteriors live in a space which has far lower dimension than the space of ``incorrect'' posteriors. In other words, the information bearing components in ``correct'' senone posteriors are fewer resulting in matrices which have lower rank compared to ``incorrect'' posteriors. Given that the ranks are nevertheless very low (compared to the dimension of the senone posteriors which is 557), the ``incorrect'' posterior are exposed to a high-dimensional spurious noise. Therefore, to enhance the posterior probabilities, 
\begin{itemize}[label={},leftmargin=*]
  \item \emph{the low-dimensional subspace has to be modeled/identified and the posterior has to be projected onto that space.}
\end{itemize}
To further investigate the subspaces selected for sparse recovery, the values in sparse representation $\alpha$ for each class are summed to form $\alpha$-sum vectors and the ``Rank'' of senone-specific $\alpha$-sum matrices are computed. According to SSR property, it is expected that sparse recovery should select the subspaces from the underlying classes so the ``Rank'' of $\alpha$-sum matrices has to be 1. In fact, we found that the empirical results averaged over the whole test set conformed to this theoretical insight indicating that 
\begin{itemize}[label={},leftmargin=*]
  \item \emph{subspace sparse recovery leads to selection of the subspaces belonging to the underlying senone classes.}
\end{itemize}
The class-specific dictionary learning for sparse coding enables us to model the non-linear manifold of the training data as a union of low-dimensional subspaces. A DNN posterior $z$ from the test data may not lie on this manifold due to presence of high-dimensional noise embedded in its components. It is important to extract the low-dimensional structure in $z$ while separating the effect of noise. Sparse coding does exactly this by finding the true underlying subspaces in sparse representation $\alpha$ and enables projecting $z$ on the class-specific subspace of the training data manifold via $\D\alpha$ reconstruction. 
\vspace{-2mm}
\subsection{Low-rank and Sparse Decomposition}\label{sec:rpca}
\begin{figure}[t]
  \centering
  \includegraphics[width=0.98\columnwidth]{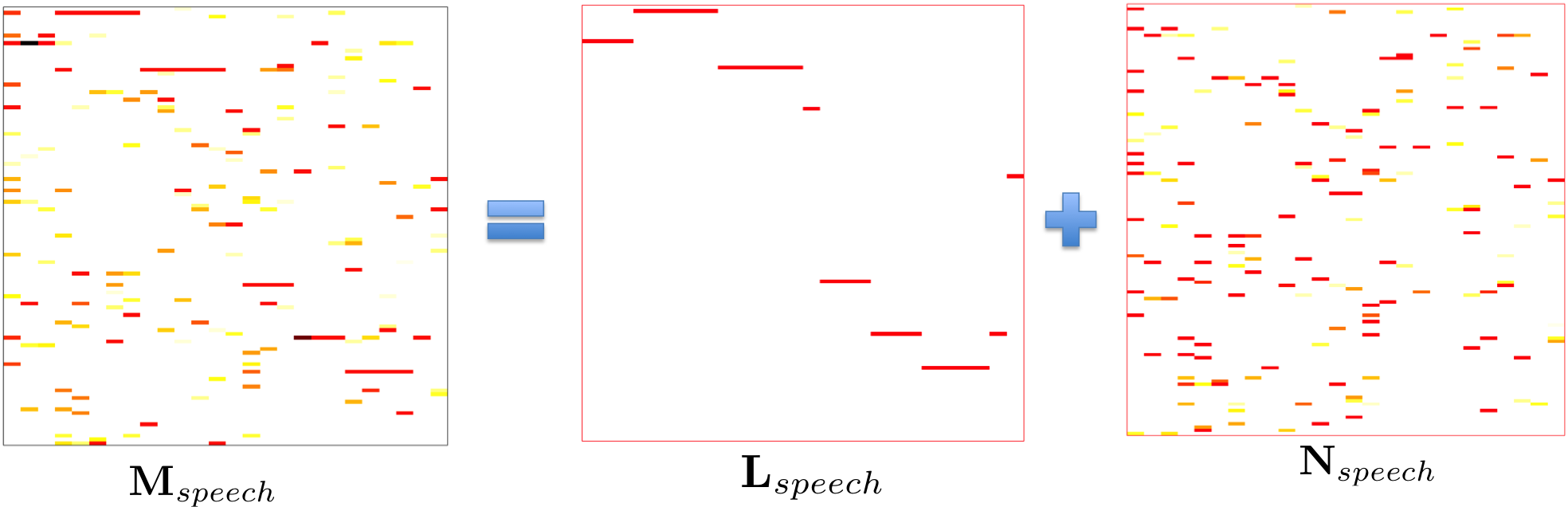}
  \vspace{-2mm}
  \caption{\footnotesize Decomposing a DNN estimated senone posterior matrix $\mathbf{M}_{\text{speech}}$ into a low-rank matrix $\mathbf{L}_{\text{speech}}$ of enhanced posteriors and a sparse matrix $\mathbf{N}_{\text{speech}}$ of spurious noise.}
  \label{fig:rpca}
\end{figure}

To further study the \emph{true} underlying dimension of the senone-specific subspaces, we consider robust principle component analysis (RPCA) based decomposition of the senone posteriors~\cite{candes2011robust}. The idea of RPCA is to decompose a data matrix $\mathbf{M}$ as
\begin{equation}\small \label{eq:rpca}
\mathbf{M}=\mathbf{L}+\mathbf{N}
\end{equation}
where matrix $\mathbf{L}$ has low-rank and matrix $\mathbf{N}$ is sparse (see Figure \ref{fig:rpca}). Building upon the observations in Section~\ref{sec:uos-reconstruct}, the low-rank component $\mathbf{L}$ corresponds to the enhanced posteriors while the high dimensional erroneous estimates are separated out in the sparse matrix $\mathbf{N}$. 

We collect posterior features for each senone from training data using \emph{ground truth} based GMM-HMM forced alignment. RPCA decomposition is applied to data of each senone-class to reveal the \emph{true} underlying dimension of the class-specific senone subspaces. The rank of senone posteriors (i.e. rank of $\mathbf{L}$) obtained after RPCA decomposition for both ``Correct'' and ``Incorrect'' classes are listed in Table~\ref{table:analysis}. We can see that the \emph{true} dimension (7.6) of the class-specific subspaces of senone posteriors is indeed far lower than the DNN posteriors (36.6) and yet lower than the projected posteriors (11.9). Exploiting this multi low-rank structure of speech can lead to posterior enhancement via low-rank representation at utterance level~\cite{liu2013robust}.  

The low-rank bottleneck layer based DNN is studied in~\cite{sainath2013low} which shows that low-dimensional structuring of DNN architecture yields smaller footprint and faster training. In contrast, our proposed method suggests an added layer of sparse coding for structuring DNN outputs relying on the generic sparse and low-rank structures. Since, these generic structures are characterized from the training data, this approach enables us to handle mismatches in DNN train and test conditions.

\vspace{-3mm}
\subsection{Enhanced DNN-HMM Speech Recognition}\label{sec:asr_results}
\vspace{-2mm}
Continuous speech recognition is performed using DNN posteriors as well as projected posteriors in the framework of conventional hybrid DNN-HMM. HMM topology learned during training of the hybrid DNN-HMM is used for decoding the word transcription in all cases. Hence, all parameters of different ASR systems shown here are the same and the only difference is in terms of senone posterior probabilities at each frame which results in different best paths being decoded by the Viterbi algorithm. 

To demonstrate the increased robustness in projected posteriors as compared to the DNN posteriors, we also compared their performance in noisy conditions where artificial white Gaussian noise was added at signal level to the test utterances at signal-to-noise (SNR) ratios of 10\,dB, 15\,dB and 20\,dB. DNN trained on clean speech is used for computing posteriors from noisy test spectral features so that the artificially added noise acts as an unseen variation in the data for DNN. Comparison of ASR performance is shown in Table~\ref{table:wer_asr} in terms of Word Error Rate (WER) percentage. 

We can see that the projected posteriors outperform DNN posteriors in all cases suggesting that projection based on $\D\alpha$ provides enhanced acoustic models for DNN-HMM decoding. We note that 
in all experiments, a consistent decrease in insertion and substitution errors is observed when using projected posteriors in place of DNN posteriors. This implies fewer wrong hypotheses being made in case of projected posteriors at word level as compared to DNN posteriors. A similar insight comes by comparing the GMM-HMM based forced senone alignment (ground truth) with senone alignments achieved by best Viterbi paths in projected posterior and DNN posterior systems. Senone classification error of 24.1\% in case of DNN posteriors is reduced to 19.8\% in case of projected posteriors. Improvement in senone alignments and subsequent reduction in WER proves superior quality of projected posteriors over DNN posteriors and supports the hypothesis that projection moves the test features closer to the subspace of the correct classes. 

Finally, RPCA posteriors (matrix $\mathbf{L}$ obtained from low-rank and sparse decomposition as explained in Section \ref{sec:rpca}) which have ranks close to the true underlying dimensions of senone subspaces perform extremely well in ASR (c.f. Table~\ref{table:wer_asr}). WER of 2.6\% using DNN posteriors (``Rank'' 36.6) reduces to a WER of 2.2\% using projected posteriors (``Rank'' 11.9) i.e. a relative improvement of 15.4\%, and when RPCA posteriors (``Rank'' 7.6) are used, it is reduced to a mere 0.4\%. Since RPCA based low-rank reconstruction of posteriors has been done using ground truth senone alignment, ASR performance in this case is the best case scenario and demonstrates the scope of improvement possible even after DNN based acoustic modeling.
{\footnotesize
\begin{table}[t]\footnotesize
\centering
\begin{tabular}{llcccc}
\hline
\textbf{SNR} & \textbf{Posteriors} & \textbf{WER (\%)} & \textbf{Ins} & \textbf{Del} & \textbf{Subs}\\
\hhline{======}
Clean  & RPCA  & 0.4 &  36 & 18 & 4 \\
\hline
\hline
\multirow{2}{*}{Clean} & DNN & 2.6 & 111 & 96 & 152\\
& Projected & 2.2 & 72 & 100 & 137\\
\hline						
\multirow{2}{*}{20db} & DNN	& 4.0	&160&	121	& 293\\
	& Projected & 3.5	&	90	&162&	233\\
\hline						
\multirow{2}{*}{15db} & DNN	&6.8	&	205&	249&	498\\
	& Projected &	6.2	&	130	&298	&442\\
\hline						
\multirow{2}{*}{10db}& DNN &14.0	&199	&950&	801\\
	& Projected &	13.9	&	117	&1064	&763\\
\hline
\end{tabular}
\vspace{-2mm}
\caption{\footnotesize Comparison of ASR performance using DNN posteriors and projected posteriors in clean and noisy conditions on Numbers'95 database. RPCA posteriors indicate an ideal enhancement through low-dimensional posterior reconstruction. Breakdown of WER in terms of insertions (Ins), deletions (Del), and substitutions (Subs) has also been shown out of a total of 13967 words in all test utterances.}
\label{table:wer_asr}
\end{table}}

 \vspace{-3mm}
\section{Conclusions and Future Directions}\label{sec:conclusions}
 \vspace{-2mm}
In this paper, we demonstrated explicit modeling of low-dimensional structures in speech using dictionary learning and sparse coding over the DNN class conditional probabilities. We showed that albeit their power in representation learning, DNN based acoustic modeling still has room for improvement in 1) exploiting the union of low-dimensional subspaces structure underlying speech data and 2) acoustic modeling in noisy conditions. Using dictionary learning and sparse coding, DNN posteriors were transformed to projected posteriors which were shown to be more suitable acoustic models. Sparse reconstruction moves the test posteriors closer to the correct underlying class of the data by exploiting the fact that the true information is embedded in a low-dimensional subspace thus separating out the high dimensional erroneous estimates. Improvements in ASR performance were shown for both clean and noisy conditions paving the way towards an effective robust ASR framework using DNN in unseen conditions. The importance of low-dimension structures was further confirmed through RPCA analysis. 

The proposed method can be improved through discriminative dictionary learning for better class-specific subspace modeling. Furthermore, we will study the low-rank clustering techniques to enhance posterior probabilities exploiting their low-dimensional multi-subspace structure. Moreover, we will consider further analysis on challenging databases and in particular the case of accented non-native speech recognition. Projection of accented speech posteriors on dictionaries trained with native language speech can result in transformation of accented phonetic space to native phonetic space and lead to improvements in accented speech recognition task. 

{
 \vspace{-2mm}
\section{Acknowledgments}
 \vspace{-2mm}
The research leading to these results has received funding from by SNSF project on ``Parsimonious Hierarchical Automatic Speech Recognition (PHASER)'' grant agreement number 200021-153507.
}

\balance
\bibliographystyle{IEEEbib}
\balance
\bibliography{strings,refs}
\end{document}